\documentclass{article}
\usepackage{ijcai25}

\usepackage{times}
\usepackage{soul}
\usepackage{url}
\usepackage[hidelinks]{hyperref}
\usepackage[utf8]{inputenc}
\usepackage[small]{caption}
\usepackage{graphicx}
\usepackage{amsmath}
\usepackage{amsthm}
\usepackage{booktabs}
\usepackage{algorithm}
\usepackage{algorithmic}
\usepackage[switch]{lineno}
\usepackage{pifont}
\usepackage{threeparttable}
\usepackage{makecell}

\usepackage{mydef}
\usepackage{hyperref}

\title{A Survey on Temporal Interaction Graph Representation Learning: Progress, Challenges, and Opportunities}

\author{
Pengfei Jiao$^1$
\and
Hongjiang Chen$^1$\and
Xuan Guo$^2$\and
Zhidong Zhao$^1$\and
Dongxiao He$^2$\And
Di Jin$^2$\thanks{Corresponding authors.}\\
\affiliations
$^1$Hangzhou Dianzi University, China\\
$^2$Tianjin University, China\\
\emails
\{pjiao, hchen, zhaozd\}@hdu.edu.cn,
\{guoxuan, hedongxiao, jindi\}@tju.edu.cn
}

\begin{document}

\maketitle

\begin{abstract}
Temporal interaction graphs (TIGs), defined by sequences of timestamped interaction events, have become ubiquitous in real-world applications due to their capability to model complex dynamic system behaviors. As a result, temporal interaction graph representation learning (TIGRL) has garnered significant attention in recent years. TIGRL aims to embed nodes in TIGs into low-dimensional representations that effectively preserve both structural and temporal information, thereby enhancing the performance of downstream tasks such as classification, prediction, and clustering within constantly evolving data environments. In this paper, we begin by introducing the foundational concepts of TIGs and emphasize the critical role of temporal dependencies. We then propose a comprehensive taxonomy of state-of-the-art TIGRL methods, systematically categorizing them based on the types of information utilized during the learning process to address the unique challenges inherent to TIGs. To facilitate further research and practical applications, we curate the source of datasets and benchmarks, providing valuable resources for empirical investigations. Finally, we examine key open challenges and explore promising research directions in TIGRL, laying the groundwork for future advancements that have the potential to shape the evolution of this field.
\end{abstract}

\vspace{-3mm}
\section{Introduction}
\begin{figure}[!t]
    \centering
    \includegraphics[width=0.95\linewidth]{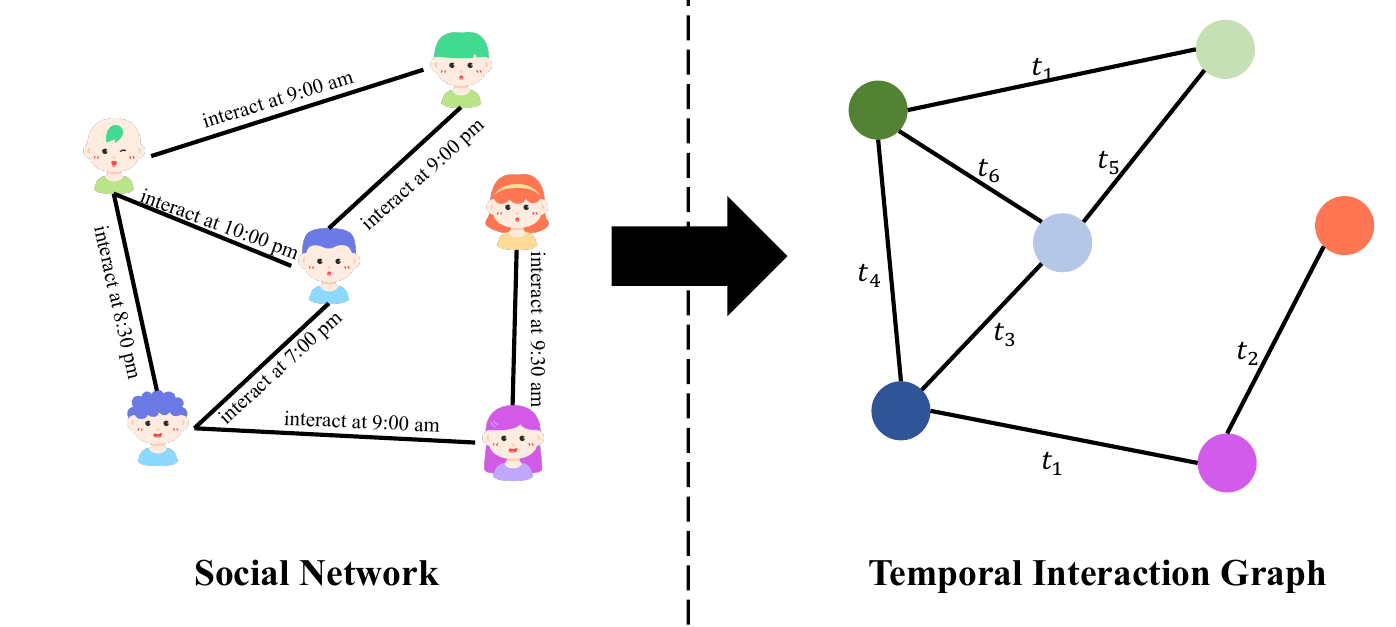}
    \caption{The overview of TIG.}
    \label{fig: example}
    \vspace{-5mm}
\end{figure}

Temporal Interaction Graphs (TIGs), also known as dynamic networks, are a powerful data structure for modeling sequences of timestamped interaction events (i.e., edges) among entities (i.e., nodes). These graphs have been widely employed to capture the evolving nature of real-world systems such as e-commerce platforms, social networks, and recommendation engines. For example, as depicted in Fig.~\ref{fig: example}, a social network can be modeled as a TIG, where nodes represent users and edges denote interactions occurring at specific timestamps. Unlike static graphs, TIGs evolve over time and can capture complex temporal dynamics and dependencies, making them indispensable in data mining and machine learning applications—including recommendation systems, chemical synthesis modeling, and social network analysis.

As the adoption of TIGs grows, the need for learning effective embeddings to enhance downstream tasks (e.g., link prediction, node classification, and node clustering) has become increasingly crucial. Traditionally, to extract useful information from a TIG, temporal data is either discarded or transformed into a sequence of static graph snapshots, which may generate potential features but often overlook fine-grained regularities in the network’s evolution. To overcome this limitation, temporal interaction graph representation learning (TIGRL) has recently attracted considerable attention. TIGRL aims to learn mappings from the input space to a low-dimensional representation while preserving both temporal dynamics and structural evolution. However, methods developed for static graphs or snapshot-based temporal graphs are not directly applicable to TIGs because of their inherently dynamic nature and irregular temporal granularity. The challenges in learning representations for TIGs are multifaceted. First, TIGs exhibit dependencies in both spatial and temporal dimensions. Their topology varies considerably over time, and the evolution patterns are often intricate and non-linear; hence, TIGRL methods must simultaneously capture static structural relationships and temporal changes. Second, many real-world systems experience rapid and unpredictable dynamics, necessitating fine-grained representations that account for both short-term fluctuations and long-term trends. In contrast, snapshot-based approaches tend to preserve temporal order while oversimplifying temporal details, thereby rendering fine-grained modeling infeasible. Moreover, these methods must also satisfy real-time reasoning requirements with computational efficiency. Finally, the domain-specific nature of TIGs introduces additional challenges: incorporating auxiliary information (e.g., node and edge attributes) often requires specialized domain knowledge, and the large-scale, continuously evolving characteristics of TIGs raise issues of scalability, storage, and parallel training.

Several surveys have investigated learning on TIGs; however, they have primarily focused on specific aspects. For example, some surveys offer a general overview of TIGs and classify models based on encoder–decoder architectures~\cite{kazemi2020representation}, while others emphasize the distinctions between TIGs and static graphs by categorizing models according to temporal modeling techniques~\cite{longa2023graph,zheng2025survey}. Additionally, certain works focus on specialized downstream tasks such as temporal link prediction~\cite{qin2023temporal} or temporal graph generation~\cite{gupta2022survey}. These efforts, while valuable, have limitations: (1) outdated coverage of recent advancements, (2) insufficient categorization of TIG methodologies, (3) limited discussion of datasets and benchmarks, and (4) inadequate attention to emerging applications and challenges.

To address these gaps, this paper provides a comprehensive review of recent advancements in TIGRL. In particular, our contributions are as follows:
\begin{itemize}
\item \textbf{A Novel Taxonomy}. We introduce a structured taxonomy that categorizes existing works based on the types of information utilized during the learning process, providing a systematic framework for navigating the field. An overview of these approaches is presented in Fig.~\ref{fig:taxonomy_of_TIGRL}.
\item \textbf{A Comprehensive Review}. Leveraging the proposed taxonomy, we conduct a thorough analysis of recent advancements in TIGRL, offering an in-depth summary of representative methods along with their respective strengths and limitations. Furthermore, we curate publicly available datasets and benchmarks to facilitate ongoing research and practical applications.
\item \textbf{Future Directions}. We highlight persisting challenges in the field and propose potential research avenues, including emerging topics such as prompt-based technologies and the integration of foundational models.
\end{itemize}

\tikzstyle{leaf4}=[draw=black, %边框
    rounded corners,minimum height=1em,
    text width=31.50em, edge=black!10, 
    %fill=hidden-orange!40,
    text opacity=1, align=center,
    fill opacity=.3,  text=black,font=\scriptsize,
    inner xsep=3pt, inner ysep=1pt,
    ]
\tikzstyle{leaf}=[draw=black, %边框
    rounded corners,minimum height=1em,
    text width=24.50em, edge=black!10, 
    %fill=hidden-orange!40,
    text opacity=1, align=center,
    fill opacity=.3,  text=black,font=\scriptsize,
    inner xsep=3pt, inner ysep=1pt,
    ]
\tikzstyle{leaf1}=[draw=black, %边框
    rounded corners,minimum height=1em,
    text width=5.28em, edge=black!10, 
    text opacity=1, align=center,
    fill opacity=.5,  text=black,font=\scriptsize,
    inner xsep=3pt, inner ysep=1pt,
    ]
\tikzstyle{leaf2}=[draw=black, %边框
    rounded corners,minimum height=1em,
    text width=6.28em, edge=black!10, 
    text opacity=1, align=center,
    fill opacity=.8,  text=black,font=\scriptsize,
    inner xsep=3pt, inner ysep=1pt,
    ]

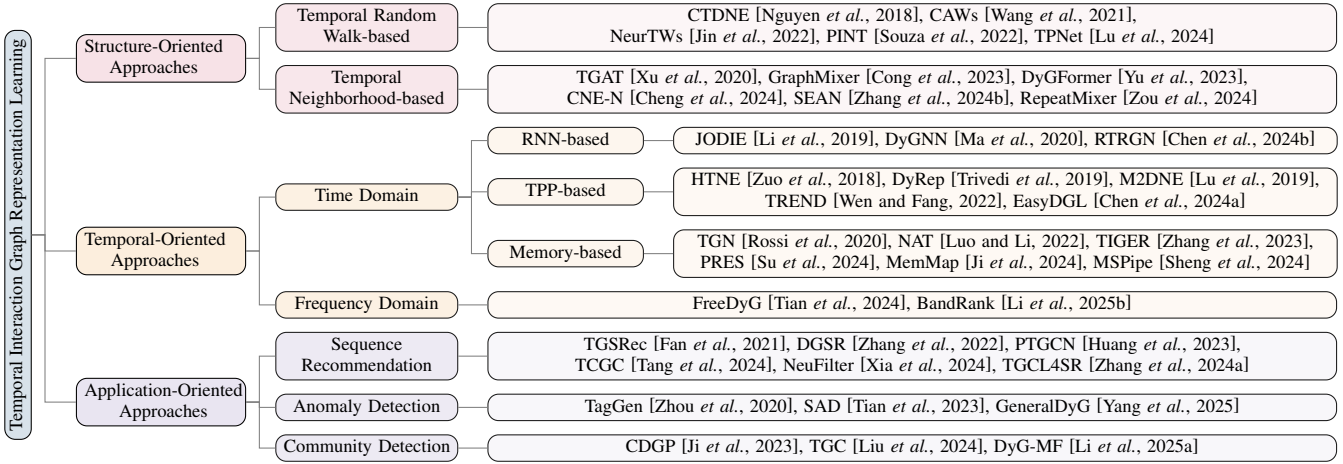
\begin{figure*}[ht]
\centering
\begin{forest}
  for tree={
  forked edges,
  grow=east,
  reversed=true,
  anchor=base west,
  parent anchor=east,
  child anchor=west,
  base=middle,
  font=\scriptsize,
  rectangle,
  draw=black, %hiddendraw 所有边框
  edge=black!50, 
  rounded corners,
  align=center,
  minimum width=2em,
  s sep=5pt,
  inner xsep=3pt,
  inner ysep=1pt
  },
  where level=1{text width=4.5em}{},
  where level=2{text width=6em,font=\scriptsize}{},
  where level=3{font=\scriptsize}{},
  where level=4{font=\scriptsize}{},
  where level=5{font=\scriptsize}{},
  [Temporal Interaction Graph Representation Learning,rotate=90,anchor=north,edge=black!50,fill=myblue,draw=black
    [Structure-Oriented \\ Approaches,edge=black!50,text width=5.8em, fill=myred
        [Temporal Random \\ Walk-based, leaf2,fill=myred
            [CTDNE \cite{nguyen2018continuous}{,} 
            % HNIP~\cite{qiu2020temporal}{,} 
            CAWs \cite{wang2021inductive}{,} \\ NeurTWs \cite{jin2022neural}{,} PINT~\cite{souza2022provably}{,} TPNet~\cite{lu2024improving}, leaf4, fill=myred]
        ]
        [Temporal \\Neighborhood-based, leaf2,fill=myred
            [  TGAT \cite{xu2020inductive}{,} GraphMixer \cite{cong2023we}{,} DyGFormer \cite{yu2023towards}{,}\\ CNE-N \cite{cheng2024co}{,} 
             SEAN \cite{zhang2024towards}{,} RepeatMixer \cite{zou2024repeat}, leaf4, fill=myred]
        ]
    ]
    [Temporal-Oriented \\ Approaches,edge=black!50,text width=5.8em, fill=myyellow
        [Time Domain, leaf2, fill=myyellow
            [RNN-based, leaf1,fill=myyellow
                 [JODIE \cite{li2019predicting}{,} DyGNN \cite{ma2020streaming}{,} RTRGN \cite{chen2024recurrent},leaf,fill=myyellow]
            ]
            [TPP-based, leaf1,fill=myyellow
                 [HTNE \cite{zuo2018embedding}{,} DyRep \cite{trivedi2019dyrep}{,} M2DNE \cite{lu2019temporal}{,} 
                 \\  TREND \cite{wen2022trend}{,} EasyDGL \cite{chen2024easydgl}, leaf,fill=myyellow]
            ]
            [Memory-based, leaf1, fill=myyellow
                [TGN \cite{rossi2020temporal}{,} NAT \cite{luo2022neighborhood}{,} TIGER \cite{zhang2023tiger}{,} 
                \\ PRES~\cite{su2024pres}{,} MemMap \cite{ji2024memmap}{,} MSPipe \cite{sheng2024mspipe}, leaf, fill=myyellow]
            ]
        ]
        [Frequency Domain, leaf2,fill=myyellow
            [FreeDyG \cite{tian2024freedyg}{,} BandRank~\cite{li2025ranking}, leaf4, fill=myyellow]
        ]
    ]
    [Application-Oriented \\ Approaches,edge=black!50,text width=5.8em, fill=mypurple
        [Sequence \\Recommendation, leaf2, fill=mypurple
            [TGSRec~\cite{fan2021continuous}{,} DGSR~\cite{zhang2022dynamic}{,} PTGCN~\cite{huang2023position}{,} \\TCGC \cite{tang2024tcgc}{,} NeuFilter~\cite{xia2024neural}{,} TGCL4SR \cite{zhang2024temporal}, leaf4, fill=mypurple]
        ]
        [Anomaly Detection, leaf2, fill=mypurple
            [TagGen~\cite{zhou2020data}{,} SAD \cite{tian2023sad}{,} GeneralDyG \cite{yang2025generalizable}, leaf4, fill=mypurple]
        ]
        [Community Detection, leaf2, fill=mypurple
            [CDGP~\cite{ji2023community}{,} TGC~\cite{liu2024deep}{,} DyG-MF~\cite{li2025revisiting}, leaf4, fill=mypurple]
        ]
    ]
  ]
\end{forest}
\caption{A taxonomy of Temporal Interaction Graph Representation Learning (TIGRL).}
\label{fig:taxonomy_of_TIGRL}
\vspace{-5mm}
\end{figure*}

\vspace{-3mm}
\section{Preliminary}
In this section, we introduce the fundamental concepts and notation used throughout the paper.

\noindent\textbf{Definitions.} A \textbf{Temporal Interaction Graph (TIG)} is a dynamic graph that models sequences of interaction events, each occurring at a specific timestamp, while also accounting for the insertion and deletion of nodes and edges. Formally, a TIG observed over a time interval $\Gamma$ is defined as a tuple $G_{\Gamma} = (V_{\Gamma}, E_{\Gamma})$, where $V_{\Gamma} = \{v_1, v_2, \dots, v_N\}$ represents the set of nodes observed during $\Gamma$. $E_{\Gamma} = \{(v_i, v_j, w_{ij}, t_e) \mid v_i, v_j \in V_{\Gamma}, \; w_{ij} \in \mathbb{R}, \; t_e \in \Gamma\}$ denotes the set of edges, where each edge $(v_i, v_j)$ is associated with a real-valued weight $w_{ij}$ and a timestamp $t_e$. This definition captures the dynamic nature of the graph’s topology, where the intervals between successive timestamps can be irregular, allowing for a detailed yet efficient representation of temporal interactions.

\noindent\textbf{Remark.} It is important to highlight a subclass of temporal graphs in which the graph structure remains static while node attributes evolve over time. These graphs, often referred to as \emph{spatio-temporal graphs}, are outside the scope of this paper. Additionally, in the literature, TIGs~\cite{chen2023temporal} are sometimes referred to as continuous-time dynamic graphs or event-based temporal graphs. TIGs can be easily converted into discrete-time dynamic graphs~\cite{jiao2024contrastive}, which we consider a special case. However, as demonstrated in~\cite{souza2022provably}, TIGs are more general and expressive, and thus will not be discussed.

\vspace{-3mm}
\section{Proposed Taxonomy}
The dynamic nature of TIGs, characterized by continuously evolving node behaviors and interaction patterns, creates intricate structural relationships and diverse temporal dependencies that complicate representation learning. To address these challenges, we introduce a taxonomy classifying existing TIGRL methods into three paradigms based on their core information utilization strategies: (1) \textbf{Structure-Oriented Approaches}, which prioritize evolving topological preservation through structural elements like node proximity, subgraph dynamics, and multi-scale motifs; (2) \textbf{Temporal-Oriented Approaches}, which model time-dependent interactions through chronological sequences, temporal event intervals, and trajectory evolution encoding; and (3) \textbf{Application-Oriented Approaches}, which adapt learning frameworks to domain-specific objectives such as recommendation systems or anomaly detection. We conduct a comprehensive methodological comparison in Table~\ref{tab: methods}, including paradigm characteristics, model techniques, task applicability, and sources.

\vspace{-1mm}
\subsection{Structure-Oriented Approaches}
Structure-oriented approaches in TIGRL focus on preserving the evolving graph topology and capturing dynamic structural patterns. These methods are designed to handle continuously changing neighborhood configurations. Broadly, current methodologies are divided into two paradigms: (1) temporal random walk-based approaches and (2) temporal neighborhood-based approaches.

\vspace{-1mm}
\subsubsection{Temporal Random Walk-Based}
Temporal random walk-based approaches capture spatiotemporal patterns by generating node sequences constrained by chronological order. A temporal random walk $W$ is defined as a sequence $W = \left[(v_0, t_0), (v_1, t_1), \dots, (v_k, t_k)\right]$, where timestamps follow certain constraints, ensuring that each transition $(v_i, v_{i+1}, t_{i+1})$ represents a valid temporal edge in the TIG. The node representation \( h_u^t \) is computed by aggregating multiple walk instances:
\begin{equation}
h_u^t = \frac{1}{M} \sum_{i=1}^M f_{\text{Enc}}(W_i),
\end{equation}
where $M$ is the number of walks, and $f_{\text{Enc}}(\cdot)$ is an encoding function implemented through sequential models like recurrent neural networks (RNNs) or transformers.
Early approaches, such as CTDNE~\cite{nguyen2018continuous}, prioritize temporal coherence during walk generation, ensuring the preservation of the chronological order of interactions. CAWs~\cite{wang2021inductive} refine this approach by incorporating causal anonymization, which enhances structural induction and captures temporal causality. NeurTWs~\cite{jin2022neural} apply neural message passing to anonymized walk sequences, while TPNet~\cite{lu2024improving} integrates temporal decay effects into the walk matrices, allowing for adaptive neighborhood weighting.

Despite their theoretical advantages in capturing temporal dependencies, these methods face practical limitations in computational efficiency due to the combinatorial explosion of valid temporal walks, particularly in dense graphs with long interaction histories. The inherent trade-off between walk length and temporal resolution further complicates parameter tuning, suggesting opportunities for future research in adaptive walk sampling strategies.
\vspace{-1mm}
\subsubsection{Temporal Neighborhood-Based}
Temporal neighborhood-based approaches derive node representations by aggregating information from dynamically evolving neighborhoods within defined time windows. The fundamental formulation for a node $u$'s embedding $h_u^t$ at time $t$ integrates weighted neighborhood features and temporal-edge attributes:  
\begin{equation}
    h_u^t = f_{\text{enc}}\left(\sum_{i \in \mathcal{N}_u^t} \alpha_{ui} \, x_i^N, \, x_i^E\right),
\end{equation}  
where $\mathcal{N}_u^t$ denotes the temporal neighborhood, $\alpha_{ui}$ adaptively weights neighbor $i$'s influence, $x_i^N$ represents node-level features or relational metrics, and $x_i^E$ encodes temporal interaction properties. The encoder $f_{\text{enc}}(\cdot)$ distills these into low-dimensional node embedding. TGAT~\cite{xu2020inductive} fuses temporal decay with attention mechanisms, DyGFormer~\cite{yu2023towards} tokenizes historical interactions for transformer-based sequence modeling, and CNE-N~\cite{cheng2024co} prioritizes co-neighbors for efficient dynamic link prediction. SEAN~\cite{zhang2024towards} and RepeatMixer~\cite{zou2024repeat} further refine $\alpha_{ui}$ via adaptive neighborhood scaling and periodic pattern detection. 

Despite excelling in temporal link prediction through neighborhood similarity modeling, these methods struggle with node-centric tasks like classification due to their relational bias and reliance on fixed time windows. This design trade-off enhances computational efficiency but limits long-term dependency modeling, balancing temporal granularity against representational complexity.

\vspace{-1mm}
\subsection{Temporal-Oriented Approaches}
Beyond topology, the temporal dimension is critical in TIGRL. Temporal-oriented approaches specifically address the challenges associated with irregular time intervals and the non-linear evolution of graph structures. Unlike conventional static graph embedding techniques that often disregard temporal variance through uniform projections, these methods embrace the irregularity of temporal signals. To achieve this, temporal-oriented approaches are further categorized into: (1) Time-Domain Approaches and (2) Frequency-Domain Approaches. % The following subsubsections detail these approaches.
\vspace{-1mm}
\subsubsection{Time Domain}
Based on the technique employed to model temporal dependencies, we categorize these approaches into three primary classes: recurrent neural network-based (RNN-based), temporal point process-based (TPP-based), and memory-based. Each category offers distinct advantages in capturing temporal patterns across diverse application scenarios.

\begin{table*}[!ht]
    \centering
    \resizebox{\linewidth}{!}{
    \begin{tabular}{clllllll}
    \toprule
    Taxonomy & Methods & Learning Paradigms & Inference Setting & Technique & Task & Sources & Code\\ \midrule
    \multirow{12}{*}{\makecell{Structure-Oriented \\ Approaches}}
    & CTDNE & Supervised Learning & Transductive & Temporal Walks & Node, Link & WWW 2018 & \href{https://github.com/Shubhranshu-Shekhar/ctdne}{link}\\
    & HNIP & Supervised Learning & Transductive & Temporal Walks & Node, Link, Graph & AAAI 2020 & -\\
    & TGAT & Supervised Learning & Inductive & GAT and GNNs & Node, Link & ICLR 2020 & \href{https://github.com/StatsDLMathsRecomSys/Inductive-representation-learning-on-temporal-graphs}{link}\\
    & CAWs & Supervised Learning & Inductive & Temporal Walks and GNNs & Link & ICLR 2021 & \href{https://github.com/KimMeen/Neural-Temporal-Walks}{link}\\
    & NeurTWs & Unsupervised Learning & Inductive & Temporal Walks and GNN & Node, Link & NeurIPS 2022 & \href{https://github.com/KimMeen/Neural-Temporal-Walks}{link}\\
    & PINT & Supervised Learning & Inductive & Temporal Walks and GNNs & Node, Link & NeurIPS 2022 & \href{https://github.com/AaltoPML/PINT}{link}\\
    & GraphMixer & Supervised Learning & Inductive & MLP and GNNs & Node, Link & ICLR 2023 & \href{https://github.com/CongWeilin/GraphMixer}{link}\\
    & DyGFormer & Supervised Learning & Inductive & Transformer and GNNs & Node, Link & NeurIPS 2023 & \href{https://github.com/yule-BUAA/DyGLib}{link}\\
    & TPNet & Supervised Learning & Inductive & Temporal Walks and GNNs & Link & NeurIPS 2024 & \href{https://github.com/lxd99/TPNet}{link}\\
    & CNE-N & Supervised Learning & Inductive & Hashtable-based Memory and GNNs & Link & SIGKDD 2024 & \href{https://github.com/ckpassenger/DyGLib_CNEN/tree/CNEN}{link}\\
    & SEAN & Supervised Learning & Inductive & RNN and GNNs & Node, Link & SIGKDD 2024 & -\\
    & RepeatMixer & Supervised Learning & Inductive & MLP and GNNs & Link & SIGKDD 2024 & \href{https://github.com/Hope-Rita/RepeatMixer}{link}\\ \midrule
    \multirow{16}{*}{\makecell{Temporal-Oriented \\ Approaches}}
    & HTNE & Supervised Learning & Transductive & Hawkes process & Node, Link, Graph & SIGKDD 2018 & -\\
    & JODIE & Supervised Learning & Inductive & GNNs and RNNs & Node, Link & SIGKDD 2019 & \href{https://snap.stanford.edu/jodie}{link}\\
    & DyREP & Supervised Learning & Inductive & TPPs and GAT & Node, Link & ICLR 2019 & \href{https://github.com/Harryi0/dyrep_torch}{link}\\
    & M2DNE & Supervised Learning & Transductive & TPPs and GNNs & Node, Link, Graph & CIKM 2019 & \href{https://github.com/rootlu/MMDNE}{link}\\
    & DyGNN & Supervised Learning & Inductive & RNN and GNNs & Node, Link & SIGIR 2020 & \href{https://github.com/alge24/dygnn}{link}\\
    & TGN & Supervised Learning & Inductive & Memory and GNNs & Node, Link & ICML 2020 & \href{https://github.com/twitter-research/tgn}{link}\\
    & NAT & Supervised Learning & Inductive & RNNs and GNNs & Node, Link & LOG 2022 & \href{https://github.com/Graph-COM/Neighborhood-Aware-Temporal-Network}{link}\\
    & TREND & Supervised Learning & Inductive & TPPs and GNNs & Node, Link & WWW 2022 & \href{https://github.com/WenZhihao666/TREND}{link}\\
    & TIGER & Supervised Learning & Inductive & Memory and GNNs & Link & WWW 2023 & \href{https://github.com/yzhang1918/www2023tiger}{link}\\
    & FreeDyG & Supervised Learning & Inductive & Frequency MLP and GNNs & Node, Link & ICLR 2024 & \href{https://github.com/Tianxzzz/FreeDyG}{link}\\
    & PRES & Supervised Learning & Inductive & Memory and GNNs & Node, Link & ICLR 2024 & \href{https://github.com/jwsu825/MDGNN_BS}{link}\\
    & RTRGN & Supervised Learning & Inductive & RNNs and GNNs & Link & NeurIPS 2024 & - \\
    & EasyDGL & Supervised Learning & Inductive & TPPs and GNNs & Node, Link & TPAMI 2024 & \href{https://github.com/cchao0116/EasyDGL}{link}\\
    & MemMap & Supervised Learning & Inductive & Memory and GNNs & Node, Link & SIGKDD 2024 & \href{https://github.com/AhaSokach/MemMap}{link}\\
    & MSPipe & Supervised Learning & Inductive & Memory and GNNs & Node, Link & SIGKDD 2024 & \href{https://github.com/PeterSH6/MSPipe}{link}\\
    & BandRank & Supervised Learning & Inductive & Band-pass Disentangle with GNNs & Node, Link & WWW 2025 & -\\ \midrule
    \multirow{13}{*}{\makecell{Application-Oriented \\ Approaches}}
    & TagGen & Supervised Learning & Inductive & Temporal Walks and GAT & Link, Detection & SIGKDD 2020 & \href{https://github.com/davidchouzdw/TagGen}{link}\\
    & APAN & Supervised Learning & Inductive & GNNs and RNNs & Node, Link & SIGMOD 2021 & \href{https://github.com/WangXuhongCN/APAN}{link}\\
    & TGSRec & Supervised Learning & Inductive & Transformer and GNNs & Recommendation & CIKM 2021 & \href{https://github.com/dygrec/tgsrec}{link}\\
    & DGSR & Supervised Learning & Inductive & Collaborative Signals with GNNs & Recommendation & TKDE 2022 & \href{https://github.com/CRIPAC-DIG/DGSR}{link}\\
    & CDGP & Unsupervised Learning & Inductive & Memory and GNNs & Node, Detection & SIGKDD 2023 & \href{https://github.com/AhaSokach/CDGP}{link}\\
    & SAD & Unsupervised Learning & Inductive & Memory and GNNs & Node, Detection & IJCAI 2023 & \href{https://github.com/D10Andy/SAD}{link}\\
    & PTGCN & Supervised Learning & Inductive & GCN and GAT & Recommendation & TOIS 2023 & \href{https://github.com/drhuangliwei/PTGCN}{link}\\
    & TCGC & Supervised Learning & Inductive & Collaborative Information with GNNs & Recommendation & TOIS 2024 & -\\
    & NeuFilter & Supervised Learning & Inductive & Kalman Filtering and GNNs & Recommendation & WSDM 2024 & \href{https://github.com/Yaveng/NeuFilter}{link}\\
    & TGCL4SR & Unsupervised Learning & Inductive & GCL and GCN & Recommendation & AAAI 2024 & -\\
    & TGC & Unsupervised Learning & Inductive & Graph Clustering with GNNs & Node, Graph, Detection & ICLR 2024 & \href{https://github.com/MGitHubL/Deep-Temporal-Graph-Clustering}{link}\\
    & GeneralDyG & Supervised Learning & Inductive & Transformer and GNNs & Node, Detection & AAAI 2025 & \href{https://github.com/YXNTU/GeneralDyG}{link}\\
    & DyG-MF & Unsupervised Learning & Inductive & Matrix Factorization and GNNs & Node, Graph, Detection & WWW 2025 & \href{https://anonymous.4open.science/r/DyG-MF}{link}\\ \bottomrule
    \end{tabular}
    }
    \caption{A summary of TIGRL models, ordered by their release time. Acronyms in Task: Node refers to node-level tasks; Link refers to link-level tasks; Graph refers to graph-level tasks; Recommendation refers to Graph Sequence Recommendation; Detection refers to Anomaly Detection or Community Detection.}
     \label{tab: methods}
     \vspace{-5mm}
\end{table*}
\noindent\textbf{RNN-based.} 
RNN-based approaches for temporal interaction graphs dynamically update node representations through sequential processing of interactions, synthesizing historical states and neighborhood context using recurrent architectures. The core mechanism can be expressed through a unified framework where a node's hidden state \( h_u^t \) evolves via:
\begin{equation}
h_u^t = \text{RNN}\left(h_u^{t^-}, \sum_{i \in \mathcal{N}_u^t} \alpha_{ui} \cdot h_i^{t^-}\right),
\end{equation}
combining a node's memory $h_u^{t^-}$ with aggregated neighborhood information from temporal neighbors $\mathcal{N}_u^t$, where attention weights $\alpha_{ui}$ modulate influence distribution. JODIE~\cite{li2019predicting} implements dual RNNs that bidirectionally update user and item embeddings, capturing mutual evolution patterns between interacting entities. RTRGN~\cite{chen2024recurrent} enhances this foundation by integrating historical neighbor states with current interactions via a Time Revision module (directly corresponding to the neighborhood aggregation term) and refining attention weights $\alpha_{ui}$ through a bias-correcting mechanism during recursive aggregation.

While these methods effectively preserve temporal dependencies through RNN mechanisms, they inherit challenges including computational demands from sequential processing and scalability limitations for large sparse graphs. 

\noindent\textbf{TPP-based.} 
TPP modeling in TIGs focuses on continuous-time interactions, using conditional intensity functions to capture both historical influence decay and spontaneous events. The Hawkes process, a variant of TPP, formalizes this dynamic through the equation:
\begin{equation}
\lambda(t) = \mu(t) + \int_{-\infty}^t \kappa(t-s) \, dn(s),
\end{equation}
where \(\mu(t)\) represents baseline interactions, and \(\kappa(t-s)\) encodes temporal dependencies between events. DyRep~\cite{trivedi2019dyrep} exemplifies a deep architectural approach by coupling dual-intensity functions with a time-aware RNN. Here, \(\lambda(t)\) governs the dynamics of interactions, while temporal attention computes \(\kappa(t-s)\) as \(\exp(-\delta(t-s))\), emphasizing recent events. HTNE~\cite{zuo2018embedding} extends this paradigm by modeling neighborhood sequences as Hawkes processes, with trainable exponential kernels that capture historical neighbor influences. TREND~\cite{wen2022trend} advances the framework by decoupling intensity into event-level and node-level components, using coupled graph neural networks (GNNs) to model both localized interactions and global trend propagation. EasyDGL~\cite{chen2024easydgl} introduces curriculum learning, adaptively weighting \(\mu(t)\) to prioritize recent events during training through TPP-based attention gates.

These methods share common strengths, such as handling asynchronous events and modeling temporal decay patterns. However, challenges remain in scaling to large graphs and addressing sparse interaction scenarios.

\noindent\textbf{Memory-based.} 
Memory-based approaches in temporal interaction graphs focus on updating node embeddings dynamically to capture temporal dependencies via event-driven memory updates. These methods follow a unified computational paradigm, consisting of three key components: message generation, memory update, and embedding computation. The process is mathematically represented as:
\begin{equation}
\begin{aligned}
 m_u^t &= \text{msg}(s_u^{t^-}, s_v^{t^-}, e_{uv}^t, t), \\
 s_u^t &= \text{mem}(s_u^{t^-}, m_v^t), \\
 h_u^t &= \text{emb}(s_u^t, \mathcal{N}_u^t),
\end{aligned}
\end{equation}
where $s_u^{t^-}, s_v^{t^-}$ are the memory states of nodes $u$ and $v$ before time $t$, \( m_i^t \) is the message generated from the event $e_{uv}^t$, $\mathcal{N}_u^t $represents the temporal neighbors of $u$ up to time $t$, and $h_i^t$ is the dynamic embedding of node $u$ at time $t$. The functions $\text{msg}(\cdot), \text{mem}(\cdot), \text{emb}(\cdot)$ are learnable modules that generate the message, update memory, and compute the embedding, respectively. TGN~\cite{rossi2020temporal} introduces RNN-based memory updates triggered by streaming events. NAT~\cite{luo2022neighborhood} improves computational efficiency with GPU-friendly dictionary structures for neighborhood representation. TIGER~\cite{zhang2023tiger} addresses long-term dependency loss by incorporating dual memory modules, while PRES~\cite{su2024pres} uses Gaussian mixture models for memory correction through time-series gradient analysis. MSPipe~\cite{sheng2024mspipe} advances parallel training by introducing controlled memory staleness, helping mitigate the temporal dependency bottleneck.

These methods excel at preserving temporal patterns through continuous memory updates and enabling real-time inference. However, they also share common challenges, including high memory storage requirements for long interaction histories, sensitivity to the event processing order, and potential error accumulation in memory states. A key design trade-off in this domain remains balancing memory fidelity with computational efficiency.

\vspace{-1mm}
\subsubsection{Frequency Domain}
The frequency domain decomposes temporal signals into constituent frequency components through mathematical transformations like Fourier analysis, providing a powerful paradigm to capture periodic patterns and long-range dependencies that are challenging to model in time-domain approaches. Building on this foundation, recent advances in TIGRL have developed innovative frequency-enhanced architectures. FreeDyG~\cite{tian2024freedyg} employs a frequency-enhanced MLP-Mixer layer that first transforms node embeddings into the frequency space via Fast Fourier Transform (FFT), then applies learnable complex-valued filters to amplify task-relevant frequencies before reconstructing time-domain features through inverse FFT. This dual-domain architecture effectively captures periodic interaction patterns and temporal shift phenomena in dynamic graphs. BandRank~\cite{li2025ranking} addresses frequency aliasing through parallel frequency-specific MLPs that extract multi-scale temporal features ranging from long-term evolutionary trends to short-term fluctuations, complemented by a harmonic ranking loss that mitigates gradient vanishing in listwise optimization. 

While these methods demonstrate superior performance in capturing cyclical patterns compared to conventional time-domain approaches, they introduce computational overhead from frequency transformations and require careful handling of phase information during inverse transforms. The frequency-domain paradigm shows particular promise for applications with inherent periodicity like social network dynamics and traffic prediction, though its effectiveness on irregular temporal patterns warrants further investigation.
\vspace{-1mm}
\subsection{Application-Oriented Approaches}
TIGRL has also been closely integrated with some specific applications, e.g., nodes, links, graphs that do not vary enough to fit a specific application. Application-oriented approaches need to be carefully considered in how to incorporate auxiliary information—such as domain knowledge or contextual attributes—to enhance representation learning for targeted tasks. Application-oriented approaches are further categorized into: (1) Sequence recommendation, (2) Anomaly Detection, and (3) Community Detection. % The following subsubsections detail these approaches.
\vspace{-1mm}
\subsubsection{Sequence Recommendation} 
The integration of TIGs into sequence recommendation systems has modeling of dynamic user-item interactions and contextual attributes. Unlike static graphs, temporal graphs encode time-stamped interactions (e.g., purchases) and auxiliary relationships (e.g., social ties), enabling fine-grained analysis of behavioral shifts, interest evolution, and activity bursts. By transforming recommendation tasks into temporal link prediction problems, methods like TIGRL effectively reduce dimensionality while preserving heterogeneous features and temporal dependencies. Recent advances address these challenges through diverse approaches: TGSRec~\cite{fan2021continuous} employs Transformer-based continuous-time representations to handle sparse, irregular data, while DGSR~\cite{zhang2022dynamic} integrates temporal attention with graph convolution for dynamic preference modeling. PTGCN~\cite{huang2023position} enhances adaptability through positional encoding and time-aware features, whereas TCGC~\cite{tang2024tcgc} jointly models interaction evolution via collaborative filtering and graph co-evolution. The TGCL4SR~\cite{zhang2024temporal} framework further advances temporal pattern extraction through contrastive learning, demonstrating robust noise resistance and multi-scale behavioral analysis. Collectively, these methods address critical challenges including data sparsity, temporal dynamics, and noise in recommendation systems.
\vspace{-1mm}
\subsubsection{Anomaly Detection} 
Anomaly Detection identifies deviations from normal behavioral patterns in TIGs by leveraging their temporal dynamics and structural evolution. Unlike static methods, TIG-based approaches exploit time-dependent node/edge variations to detect subtle anomalies like fraudulent transactions in financial networks or malicious activities in social platforms. A common framework integrates temporal graph encoders with anomaly scoring modules, where node representations are dynamically updated to reflect evolving interactions, and deviations are measured through reconstruction errors or outlier scoring. SAD~\cite{tian2023sad} advances semi-supervised detection via contrastive learning on temporal subgraphs, reducing dependency on labeled anomalies. GeneralDyG~\cite{yang2025generalizable} enhances generalizability through temporal ego-graph sampling and hybrid GNN-Transformer architectures, addressing feature diversity across domains. While these methods excel in capturing temporal irregularities, they often face challenges in balancing detection sensitivity with computational efficiency, particularly in streaming scenarios. Key limitations include limited robustness to adversarial temporal perturbations and reliance on predefined temporal windows. 
\vspace{-1mm}
\subsubsection{Community Detection}
Community detection in TIGs analyzes topological patterns and temporal dependencies to uncover evolving network structures, capturing dynamic behaviors like community splitting, merging, and membership migration through continuous interaction updates. Unlike static methods, TIG-based approaches enable incremental updates that balance efficiency with temporal coherence while adapting to structural shifts. Recent technical advances employ diverse strategies: CDGP~\cite{ji2023community} dynamically integrates temporal-structural signals via graph propagation for evolution tracking, while TGC~\cite{liu2024deep} adjusts deep clustering techniques to suit the interaction sequence-based batch-processing pattern of TIG. DyG-MF~\cite{li2025revisiting} reformulates dynamic matrix factorization using temporal latent factors to detect gradual community changes. Despite progress, these methods struggle with abrupt structural shifts due to temporal pattern over smoothing in long sequences, computational bottlenecks from continuous updates, and sensitivity to interaction sparsity. Current solutions partially address these through hybrid architectures like residual compression and multi-task learning frameworks, yet fundamental trade-offs persist between temporal resolution and scalability.

\vspace{-3mm}
\section{Frontier Technology for TIG}
This section presents frontier technologies for TIG, including generative learning, prompt learning, and foundational models, which are pivotal advancements shaping the future of TIGRL.
\vspace{-1mm}
\subsection{Generative Learning}
Generative learning for TIGs focuses on modeling data distributions to synthesize instances preserving both structural and temporal dynamics inherent in evolving entity interactions. Traditional frameworks like VAEs, GANs, and diffusion models have been adapted to TIGs by integrating temporal mechanisms such as TPP and RNN, enabling the generation of chronologically valid interaction sequences with accurate topological patterns. Recent advancements employ hybrid architectures to address the interplay between temporal coherence and structural fidelity. For example, TG-GAN~\cite{zhang2021tg} utilizes a continuous-time GAN with recurrent generators to produce timestamped edge sequences while maintaining chronological constraints, whereas TIGGER~\cite{gupta2022tigger} combines TPPs with autoregressive decoders to capture interaction bursts and long-term dependencies  MTM~\cite{liu2023using} introduces higher-order structural motifs (e.g., triadic closures) to guide generation, preserving local-global dynamics through transformer-based temporal embeddings. These models successfully replicate critical metrics like degree distributions, temporal motif frequencies, and inter-event intervals in real-world datasets, demonstrating their capacity to generate realistic TIGs despite challenges in modeling irregularly sampled or continuous-time interactions.
\vspace{-1mm}
\subsection{Prompt Learning}
Prompt learning, initially rooted in natural language processing for adapting pre-trained models via task-specific prompts without modifying core parameters, has been extended to TIGs to address dynamic and time-sensitive relational patterns. Unlike static graphs, TIGs require prompts to encapsulate structural dependencies and temporal dynamics, such as evolving user-item interactions in e-commerce or social network communications over time. Recent advancements integrate temporal encoding mechanisms—like dynamic context-aware templates and node-time conditional prompts—to align pre-training objectives with time-dependent tasks. For example, DyGPrompt~\cite{yu2024dygprompt} introduces prompts conditioned on node attributes and timestamps, dynamically adjusting temporal neighborhood focus during message propagation, while TIGPrompt~\cite{chen2024prompt} employs task-specific tokens to modulate attention to temporal granularity (e.g., short-term vs. long-term dependencies) through joint learning with positional encodings. By freezing backbone networks and tuning lightweight prompts, these methods achieve efficiency and scalability for large-scale TIGs with continuous temporal streams, balancing adaptability with computational constraints. This paradigm shift underscores the potential of temporal prompt engineering in bridging static model architectures with dynamic real-world interaction patterns, offering insights for both graph mining and broader AI communities.
\vspace{-1mm}
\subsection{Foundation Model}
Foundation models have transformed machine learning by learning transferable representations through large-scale pre-training, with emerging applications in TIGs that model evolving relationships in domains like social networks and financial transactions. These models unify temporal dependencies, structural evolution, and contextual patterns into a universal framework through self-supervised learning on massive dynamic graph datasets. While traditional approaches rely on task-specific architectures, recent TIG-oriented foundation models integrate temporal-aware attention mechanisms, continuous-time encoding, and multi-scale aggregation to handle heterogeneous scenarios ranging from recommendation systems to epidemic prediction. A key innovation involves adapting large language models (LLMs) for temporal graph reasoning: methods like LLM4DyG~\cite{zhang2024llm4dyg} encode temporal interactions and node attributes into text prompts, enabling LLMs to predict future states via text-generation paradigms. Hybrid architectures also combine pre-trained temporal graph encoders with task-specific adapters, using techniques like time-decayed attention to prioritize recent interactions while maintaining long-term dependencies. However, most current solutions repurpose existing frameworks like transformers rather than building architectures with native support for TIG inductive biases, creating tension between computational efficiency and the inherent complexity of temporal-structural patterns in TIGs.

\vspace{-3mm}
\section{Datasets and Benchmarks}
This section introduces commonly used datasets and benchmarks in TIGRL, which are crucial for validating and comparing algorithms to drive research progress in this field.
\begin{table}[!ht]
    \centering
    \resizebox{\linewidth}{!}{
    \begin{tabular}{ccccccc}
    \toprule
    Datasets & Domain & $|V|$ & $|E|$ & $|T|$ & Sizes & Sources\\
    \midrule
    Wikipedia & Social & 9,227 & 157,474 & 152,757 & 534MB & \href{https://zenodo.org/records/7213796\#Y1cO6y8r30o}{link}\\  
    Reddit & Social & 10,984 & 672,447 & 669,065 & 2.2GB & \href{https://zenodo.org/records/7213796\#Y1cO6y8r30o}{link}\\  
    Enron & Social & 184 & 125,235 & 22,632 & 35MB & \href{https://zenodo.org/records/7213796\#Y1cO6y8r30o}{link}\\ 
    UCI & Social & 899 & 33,720 & 58,911 & 668KB & \href{https://zenodo.org/records/7213796\#Y1cO6y8r30o}{link}\\ 
    ML25M & Social & 221,588 & 25,000,095 & 866,346 & 647MB & \href{https://grouplens.org/datasets/movielens/25m}{link}\\  
    GDELT & Social & 16,682 & 191,290,882 & 170,522 & 82.3GB & \href{https://github.com/amazon-science/tgl/blob/main/down.sh}{link}\\  
    MOOC & Interaction & 7,144 & 411,749 & 345,600 & 40MB & \href{https://zenodo.org/records/7213796\#Y1cO6y8r30o}{link}\\ 
    LastFM & Interaction & 1,980 & 1,293,103 & 1,283,614 & 37MB & \href{https://zenodo.org/records/7213796\#Y1cO6y8r30o}{link}\\ 
    Social Evolution & Proximity & 74 & 2,099,520 & 565,932 & 148MB & \href{https://zenodo.org/records/7213796\#Y1cO6y8r30o}{link}\\ 
    Contact & Proximity & 692 & 2,426,279 & 8,065 & 160MB & \href{https://zenodo.org/records/7213796\#Y1cO6y8r30o}{link}\\ 
    Flights & Transport & 13,169 & 1,927,145 & 122 & 32MB & \href{https://zenodo.org/records/7213796\#Y1cO6y8r30o}{link}\\ 
    Bitcoin-OTC & Economics & 5,881 & 35,592 & 35,592 & 988KB & \href{https://snap.stanford.edu/data/soc-sign-bitcoin-otc.html}{link}\\ 
    Bitcoin-Alpha & Economics & 3,782 & 24,186 & 24,186 & 492KB & \href{https://snap.stanford.edu/data/soc-sign-bitcoin-alpha.html}{link}\\ 
    DGraphFin & Economics & 4,889,537 & 4,300,999 & 1,433 & 649MB & \href{https://dgraph.xinye.com/dataset}{link}\\ 
    \bottomrule
\end{tabular}}
\caption{Dataset statistics}
\label{tab: datasets}
\vspace{-4mm}
\end{table}

\vspace{-2mm}
\subsection{Datasets}

High-quality datasets form the cornerstone of TIG research by enabling rigorous algorithm validation and comparative analysis across diverse scenarios. As summarized in Table~\ref{tab: datasets}, publicly available TIG datasets exhibit significant variations in scale, ranging from compact networks like UCI (thousands of nodes/edges) to massive real-world systems such as ML25M (millions of interactions). This scalability spectrum supports both granular pattern discovery and large-scale industrial applications. The datasets span multiple domains including social dynamics, transportation flows, and economic transactions, demonstrating TIG's broad applicability in modeling temporal relational systems. Notably, these resources primarily capture structural evolution patterns while lacking rich node/link attributes—a limitation stemming from practical data collection challenges. This constraint paradoxically encourages research into generalizable models that infer latent features from temporal topology changes, potentially enhancing robustness across application scenarios.

\begin{table}[!ht]
\centering
\resizebox{\linewidth}{!}{
\begin{tabular}{lll}
\toprule
\textbf{Benchmark} & \textbf{Code} & \textbf{Year}\\
\midrule
PyTorch Geometric Temporal & \url{https://github.com/benedekrozemberczki/pytorch_geometric_temporal} & 2021\\
TGL & \url{https://github.com/amazon-science/tgl} & 2022\\
Dynamic Graph Library & \url{https://github.com/yule-BUAA/DyGLib} & 2023\\
Temporal Graph Benchmark & \url{https://github.com/shenyangHuang/TGB} & 2024\\
BenchTemp & \url{https://github.com/qianghuangwhu/benchtemp} & 2024 \\
Dynamic Graph Benchmark& \url{https://github.com/gravins/dynamic_graph_benchmark} & 2024 \\
BenchTGNN & \url{https://github.com/Yang-yuxin/BenchTGNN} & 2024 \\
\bottomrule
\end{tabular}}
\caption{Links of benchmarks}
\label{tab: benchmarks}
\vspace{-4mm}
\end{table}

\vspace{-2mm}
\subsection{Benchmarks}
Open-source benchmarks are essential for advancing temporal interaction graph research by providing standardized evaluation protocols that facilitate reproducible research and systematic performance comparisons. Table~\ref{tab: benchmarks} catalogs several key resources: PyTorch Geometric Temporal offers lightweight temporal GNN implementations optimized for small-scale prototypes, while TGL overcomes scalability challenges with distributed training architectures for billion-edge graphs. The Temporal Graph Benchmark establishes multi-domain evaluation standards across more than ten datasets with rigorous temporal integrity checks, BenchTGNN reveals critical design patterns in temporal message passing and neighborhood sampling through ablation studies, and the Dynamic Graph Benchmark focuses on industrial-grade metrics to monitor concept drift robustness in streaming scenarios; collectively, these benchmarks address diverse aspects of temporal graph learning and provide a comprehensive toolkit for the field.

\vspace{-3mm}
\section{Conclusion and Future Directions}
In this survey, we provide a comprehensive overview of recent advancements in TIGRL. We introduce a novel taxonomy that categorizes existing research from multiple perspectives. Our analysis reveals that the majority of current efforts focus on addressing node correlation, capturing temporal patterns, enhancing robustness to noise, and scaling to large-scale networks. Despite these significant advancements, TIGRL faces several challenges and presents a variety of promising directions for future research:

\noindent \textbf{More Complex TIGs.}
The development of TIG learning is constrained by critical gaps in dataset construction and evaluation practices. While recent advances have produced large-scale dynamic graph datasets, their structural and complexity lag far behind static counterparts, as evidenced by EdgeBank's~\cite {poursafaei2022towards} competitive performance only using simple historical interaction patterns. This highlights the inadequacy of existing benchmarks in distinguishing sophisticated models, necessitating datasets with multi-scale hierarchies, realistic evolution dynamics (e.g., edge deletions, attribute shifts), and specialized graph types like temporal signed networks and heterogeneous graphs for applications such as financial fraud detection. Current repositories often neglect crucial real-world temporal operations and multidimensional features, limiting the comprehensive evaluation of model robustness. The research community needs to prioritize creating annotated TIG benchmarks that integrate heterogeneous interaction modalities, temporal granularity, and domain-specific topological patterns. Such efforts will enable rigorous validation of spatiotemporal dependency modeling and advance the practical deployment of TIG models across industries requiring dynamic relational reasoning.

\noindent \textbf{Explainability and Interpretability.}
Explainability and interpretability are critical for enhancing trust and transparency in TIG models, particularly in dynamic, time-sensitive domains such as finance, healthcare, and social networks. These properties aim to clarify how models infer temporal patterns, attribute importance to interactions, and rationalize predictions in evolving graph structures. Recent studies have proposed tailored methods to address TIG-specific challenges. For instance, the T-GNNExplainer~\cite{xia2023explaining} decomposes temporal reasoning into exploration (identifying influential temporal paths) and navigation (highlighting key interactions), enabling stepwise explanations for dynamic graph predictions. Another approach, TempME~\cite{chen2023tempme}, leverages motif discovery to uncover recurring temporal subgraph patterns that drive model decisions, bridging low-level interactions and high-level predictions. These studies show that although the interpretability of TIGRL is a relatively new and less discussed area, there are already some studies attempting to address this issue.

\noindent \textbf{Unified Evaluation Standards.}
While benchmarks like TGL and DyGLib have advanced TIG evaluation, current standards suffer from critical limitations in protocol design and task adaptability. The oversimplified evaluation for link prediction tasks, though achieving high performance on specific datasets, inadequately reflects real-world complexity where temporal interactions require multi-class classification, regression, or anomaly detection capabilities. This discrepancy stems from insufficient standardization in handling non-classification tasks, particularly the absence of unified normalization for continuous interaction intensities and inconsistent negative sampling strategies for anomaly detection. To enable fair cross-model comparisons, future frameworks must implement multi-dimensional metrics incorporating temporal dynamics and interaction intensity. Crucially, this requires establishing unified preprocessing pipelines, dynamic evaluation protocols simulating real-world streaming scenarios, and domain-specific benchmarks mirroring practical deployment conditions.

\section*{Acknowledgments}
This work was supported in part by the Zhejiang Provincial Natural Science Foundation of China under Grant LDT23F01012F01, in part by the National Natural Science Foundation of China under Grants 62372146, 62422210, and 92370111.

\bibliographystyle{named}
\bibliography{ijcai25}

\end{document}